\documentclass[journal]{IEEEtran}
\ifCLASSINFOpdf
\else
\fi

\usepackage{microtype}
\usepackage{graphicx}
\usepackage{subfigure}
\usepackage{booktabs} 
\usepackage{multirow}
\usepackage{bbding}
\usepackage{float}
\usepackage{algorithm}
\usepackage{algorithmic}
\usepackage{cite}
\begin{document}
%
\title{Proposal Refinement for Few-Shot Object Detection}
%
%
%

\author{Yuan~ Zeng,
        Bin~Song,~\IEEEmembership{Senior Member,~IEEE,}
        Jie~Guo,~\IEEEmembership{Member,~IEEE,}
        and~Yuwen~Chen
\thanks{$(Corresponding~author: Bin~Song.)$
	
	Bin Song, Yuan Zeng, Jie Guo and Yuwen Chen are with State Key Laboratory of Integrated Services Networks, Xidian University, 710071, China, e-mail: bsong@mail.xidian.edu.cn; zengyuan-flyvideo@outlook.com; jguo@xidian.edu.cn; chywhg@gmail.com}}

%
%

\markboth{Journal of \LaTeX\ Class Files,~Vol.~14, No.~8, August~2015}%
{Shell \MakeLowercase{\textit{et al.}}: Bare Demo of IEEEtran.cls for IEEE Journals}
%



\maketitle

\begin{abstract}
Few-shot object detection has gained widely attention in recent years. Some excellent algorithms have been proposed to handle this task. However, most of these algorithms rely on the performance of few-shot classification. Unlike previous attempts, our work focuses on the problem of unbalanced distribution of region proposals between the novel classes and the base classes. In order to alleviate this unbalanced distribution, we propose the proposal refinement approach for different training phases. Specifically, refinement loss is designed for the base training phase to enhance sensitivity of the model to novel classes, and refinement branch is introduced as an auxiliary branch for RPN (Region Proposal Networks) to generate more novel proposals in the fine-tuning phase. By rebalancing the proposal distribution, the proposed approach outperforms the baselines methods by roughly 1\%$\sim$6\% on current benchmarks without increasing any inference time. Through extensive experiments, we prove that we establish a new state-of-the-art method for the few-shot object detection task.
\end{abstract}


%
\IEEEpeerreviewmaketitle

\section{Introduction}
%
%
%
%
\IEEEPARstart{W}{ith} the help of deep learning, the object detection technology \cite{Faster17,yolov217,centernet} has grown by leaps and bounds. However, these achievements all rely on the precondition of a large number of well-labeled samples. For most pattern recognition related tasks, data labeling is a quite time-consuming work and the samples of some categories are very hard to collect, such as rare wildlife, high risk industrial defects and so on. When the training data is scarce, deep neural network seems will fail to converge or fall into the overfitting dilemma. Therefore, few-shot learning appears to try to solve the problem of insufficient training data. 

Meta-learning\cite{MAML,L2C,Intelligence} and metric learning\cite{matching,Prototypical,D2N4} have achieved SOTA (state-of-the-art) results on the field of image classification under few-shot setting. However, due to the complexity of the object detection paradigm, only a small number of few-shot object detection algorithms have been proposed. For example, FSRW\cite{FSRW} propose a pipeline which combines one-stage object detection method with meta-learning, and RepMet\cite{RepMet} takes advantage of metric learning to improve the classification performance of the network.\cite{fsdet}  design a two-stage fine-tuning approach (TFA) based on transfer learning outperforms most previous state-of-the-art methods by a large margin on the existing PASCAL VOC\cite{VOC} and COCO\cite{COCO} benchmarks. 

Comparing with methods based on meta-learning and metric learning, TFA has shown its effectiveness and intuitiveness in few-shot object detection task. The TFA is also widely applied in other research fields, such as long-tail object detection task\cite{group_softmax,Decoupling,GUO202187} and few-shot image classification task\cite{Self_Supervised}. But the feature extraction module training by TFA is insensitive to the feature of novel classes, which directly leads to the disadvantage on the number of novel proposals compared to base proposal during the fine-tuning phase. As shown in Figure 1, the number of base proposals surpasses novel proposals by a large margin under different few-shot settings. The accuracy of two-stage detectors highly relies on the quantity and quality of proposals, therefore, the unbalanced distribution of proposals lead to the difficulty of improving the detection precision of novel classes.

\begin{figure*}[ht]
	\vskip 0.1in
	\begin{center}
		\centerline{\includegraphics[width=1\textwidth]{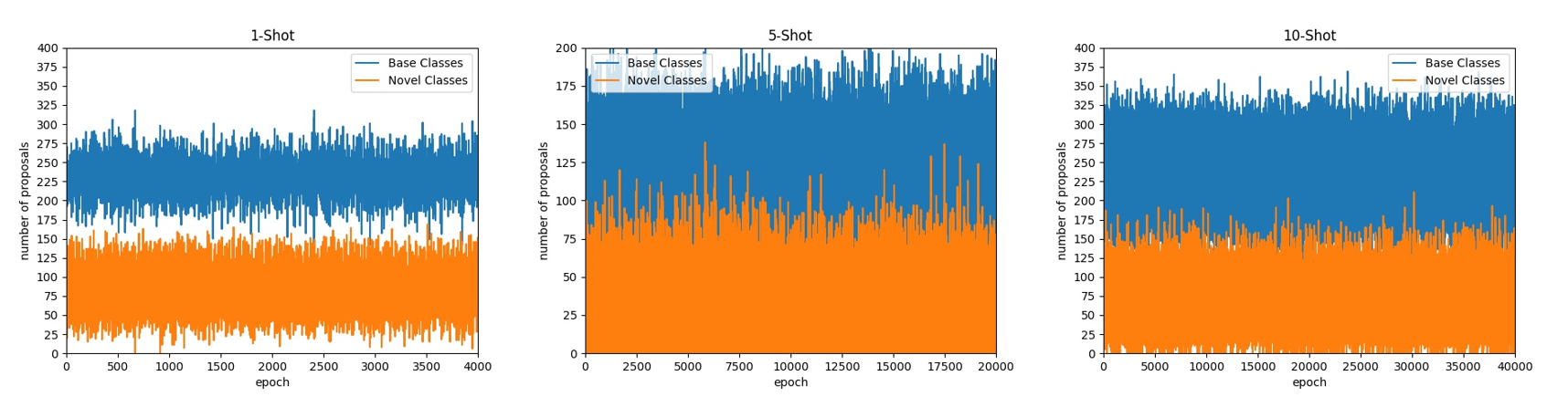}}
		\caption{Comparison of base proposal and novel proposal in quantity during the fine-tuning phase under different few-shot setting.}
		\label{fig1}
	\end{center}
	\vskip 0.1in
\end{figure*}

In this work, we introduce the proposal refinement approach for different training phases of few-shot object detection, aiming at alleviating the imbalance of proposal distribution between novel classes and base classes. We take the reputed Faster R-CNN with FPN\cite{FPN} as the basic detection model. And the TFA is utilized as our base training strategy. During the base training phase, in order to improve the feature extraction ability for novel classes, we use all the samples including base classes and novel classes to train the model. However, when a sample of a certain class is utilized for training, the parameters of the other classes will receive discouraging gradients which lead the other classes to predict low probabilities. Since the sample of base classes occupies an overwhelming advantage in data distribution, the parameters of novel classes will receive a large amount of discouraging gradients from base classes. So, the network could hardly learn any feature of novel classes. This is the reason why previous few-shot object detection methods only use the data of base classes in the base training phase. In order to enhance the sensitivity of the feature extraction network to the features of novel classes and balance the number of proposals between novel classes and base classes in the beginning of the fine-tuning phase, we use all the samples belonging to base classes and novel classes to training the model in the base training phase and the refinement loss is proposed to alleviate the negative impact of discouraging gradients generated by base classes on novel classes learning.

In the fine-tuning phase, we propose the refinement branch as an auxiliary branch for RPN to help the selection of novel proposals. There are two branches in original RPN: objectness branch to determine whether the anchor is positive or negative, and the regression branch to regress the relative offset of anchors to the ground truth boxes. The original RPN would select region proposal boxes from a large quantity of anchors according to the score generated by objectness branch. We propose refinement branch as the third branch of RPN, which could be used to decide whether the proposal belongs to novel classes. With the help of refinement branch, the RPN could select proposals according to mixed score of objectness branch and refinement branch and more proposals of novel classes could be generated and the number of base proposals will also be suppressed.

Our key contributions can be summarized as follows:
\begin{itemize}
	\item We analyze the few-shot object detection problem from a novel perspective: the unbalanced proposal distribution between novel classes and base classes caused by the unbalanced data distribution. 
	\item Our proposed refinement loss and refinement branch could alleviate the adverse effect of the unbalanced proposal distribution without increasing any inference time.
	\item We present extensive experiments over different datasets, and our proposed few-shot detector outperforms baseline methods in a variety of datasets and different settings. 
\end{itemize}

\section{Related Work}

\subsection{Object detection}
The DNN (deep neural network) based object detector could be roughly divided into two categories: one-stage detector\cite{Cascade, Mask, Fast} and two-stage detector\cite{CornerNet, focalloss, yolov1}. Specifically, one-stage object detectors focus on the trade-off between precision and inference time. In recent years, one-stage object detectors gradually developed into two branches further: anchor-based detector\cite{focalloss,yolov217} and anchor-free detector\cite{centernet,CornerNet}. The representative anchor-based method is YOLOv3\cite{YOLOv3}, which exhaust anchors centered on each pixel of the feature map, and then regress all the required information on those anchors, such as categories, offsets, and box sizes. The representative anchor-free method is CenterNet\cite{centernet}. This method regresses center point and category of targets on the feature map directly. Different from one-stage detectors, most two-stage object detectors are proposal-based, and these detectors are generally designed with an independent network to generate region proposal boxes. The representative methods are the RCNN\cite{RCNN} series algorithms. The most recognized methods among them is Faster R-CNN, which adopts region proposal network to generate a large number of candidate proposals on feature map in the first step, then regress the category, size and offset of these candidate proposals.

\subsection{Few-shot classification}
The few-shot classification methods can be roughly divided into three branches:
\textbf{(a)} Meta-learning-based methods aim to quickly update network parameters by utilizing the task-level meta knowledge acquired by a carefully designed architecture.
The most recognized few-shot classification method based on meta-learning is proposed by Finn et al.\cite{MAML}, which a training method lead the model have good performance on new few-shot tasks with only a few gradient updates.
\textbf{(b)} Metric-learning-based methods try to help the model learns a more discriminative metric space, which could increase the distance between different classes. The idea is similar to the application of nearest neighbor algorithm on classification task. In particular, Relation Network\cite{L2C} learns an embedding space and a deep non-linear distance metric to compare support and query items. Matching network\cite{matching} learns a network that maps the target example to the labelled support set. Snell et al.\cite{Prototypical} propose Prototypical Networks which could be considered as an upgrade of Matching Networks, linear classifier is replaced with weighted nearest neighbor classifier for each class. 
\textbf{(c)} Domain adaptation is a novel direction for solving the few-shot classification problem. The feature distribution of the novel classes is different from that of those base classes, resulting in poor generalization even when a model is meta-trained on the base classes. Guan et al.\cite{DAPNA} propose a domain adaptation prototypical network with attention (DAPNA) to explicitly tackle such a domain shift problem in a meta-learning framework.

\subsection{Few-shot object detection}
Few-shot object detection is a field of great practical value, and researches in the field are still very cutting-edge and worth exploring. Based on transfer learning, LSTD\cite{LSTD} utilizes the regularization technique to alleviate overfitting problem under few-shot situation. RepMet\cite{RepMet} applied distance metric learning to object classification under few-shot setting, this method learns the backbone network parameters, the embedding space, and the multi-modal distribution in a single end-to-end training process. Kang et al.\cite{FSRW} reconstructed the object detection paradigm with the idea of meta-learning. This method reweights the feature of query set with the meta knowledge of support set, and the reweighted features could detect novel objects with only a few samples. Similarly, Meta RCNN\cite{MetaRCNN} delivers a two-stage detection architecture and proposing meta-learning over RoI (Region-of-Interest) features instead of a full image feature. Juan-Manuel et al.\cite{ONCE} have investigated the challenging yet practical incremental few-shot object detection problem, the proposed method ONCE is capable of incrementally registering novel classes with few examples in a feed-forward manner, without revisiting the base class training data. Fan et al.\cite{FSOD} introduced  Attention-RPN, Multi-Relation Detector and Contrastive Training strategy in few-shot object detection method, which exploit the similarity between the few shot support set and query set to detect novel objects while suppressing false detection in the background. To handle the the problem of scale variations under few-shot setting, MPSR\cite{MPSR} generates multi-scale positive samples as object pyramids and refines the prediction at various scales. FsDet\cite{fsdet} solve the problem in an intuitive and effective way by proposing a two-stage fine-tuning approach (TFA) based on transfer learning. 

Our work is motivated by the research of the FsDet. Based on FsDet, we make further researches on the unbalanced proposal distribution between novel classes and base classes and introduce a proposal refinement approach which could alleviate this situation.

\section{Methodology}
In this section, we first introduce our training strategy under few-shot object detection setting. Then we talk about the details of our proposed method in Section 3.1 and Section 3.2. We choose FsDet as our baseline method and adopt the widely used Faster R-CNN\cite{Faster17}, a two-stage object detector, as our base detection model. Generally speaking, there are three modules in Faster R-CNN detection paradigm: Backbone for extracting features, RPN for generating region proposals and RoI (region of interest) Heads for the classification and regression of proposals. RoI Heads can be further divided into four components: RoI Pooling for aligning proposal features, RoI Feature Extractor for proposal level feature extraction, Box Classifier to classify the object categories and Box Regressor to predict the bounding box coordinates. 

\begin{figure}[h]
	\vskip 0.1in
	\begin{center}
		\centerline{\includegraphics[width=0.53\textwidth]{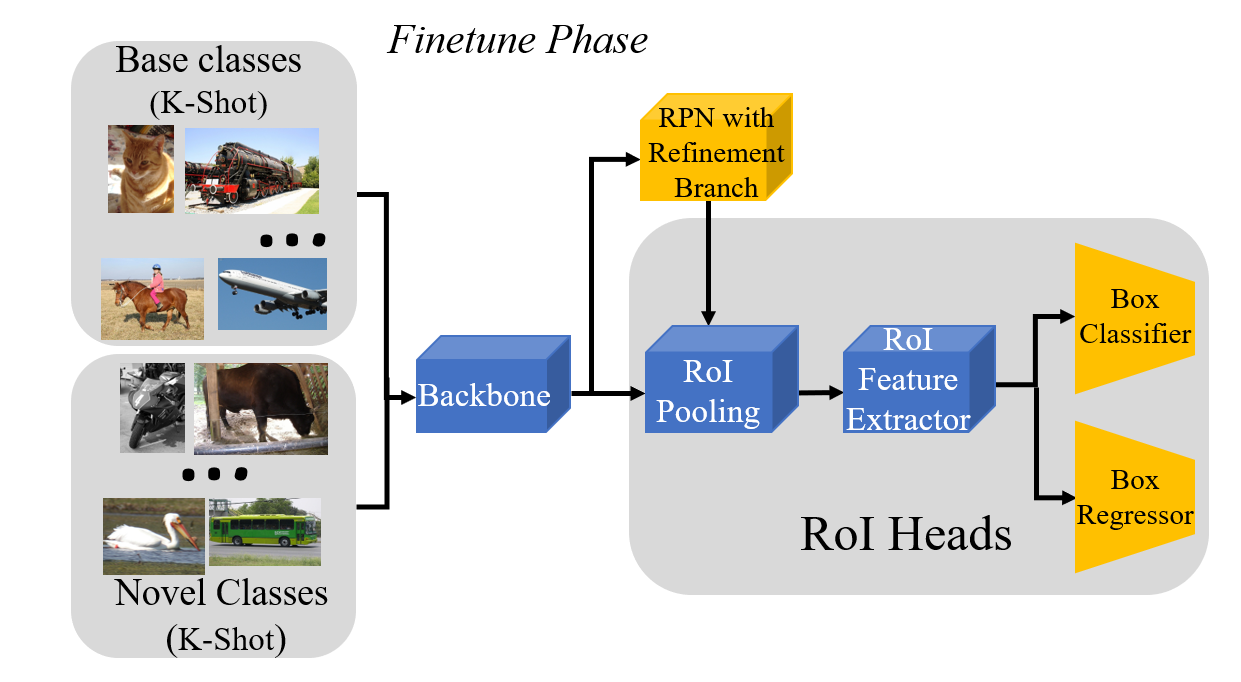}}
		\caption{ The training pipeline during the fine-tuning phase. The parameters of modules in bule are fixed during fine-tuning phase, and the modules in yellow update their parameters with SGD (Stochastic Gradient Descent).}
		\label{fig2}
	\end{center}
	\vskip 0.1in
\end{figure}

During base training phase, we train our model with all the samples in the dataset like original Faster RCNN training strategy. And we replace the softmax cross-entropy loss function of Box Classifier with our proposed refinement loss function. In the fine-tuning phase, as shown in Figure 2, the training dataset is a small balanced training set with K shots per class, which contains all classes following the setting of the baseline method. Layers corresponding to novel classes of Box Classifier and Box Regressor are randomly initialized before fine-tuning phase. All the modules except RPN, Box Classifier and Box Regressor are frozen during the fine-tuning phase.

\subsection{Refinement Loss} Wang et al.\cite{fsdet} propose to train the network with only samples of base classes in the base training phase and freeze the backbone and proposal generator during the fine-tuning phase. This training strategy could achieve promising results in a very efficient way. However, the feature extractor and proposal generator trained by this training strategy are insensitive to the contour feature, semantic feature and texture feature of novel classes, thus the Box Classifier and Box Regressor could not receive high quality proposal boxes from the RPN network. So, it is intuitive to train the network with all the samples of novel classes and base classes during the base training phase. 

Since the quantity of instances of base classes have overwhelming advantage over novel classes under few-shot setting, the network mainly performs gradient backpropagation based on the loss calculated by the base instances (instances belong to base classes) and its ground truth boxes. When a foreground sample which belongs to base classes performs gradient backpropagation, other categories will receive discouraging gradients for model updating, which might cause low probability for other categories. Base classes which have enough foreground samples could ignore the influence of discouraging gradient. But the accumulated discouraging gradients from other categories will have a non-negligible impact on novel classes. Finally, even positive samples of novel classes would get a relatively low probability from the network.

To solve this problem, we refer to the Equalization Loss\cite{EQloss} in long-tail object detection task and propose the refinement loss for the Box Classifier of the RoI Heads. Formally, we introduce reweighting factor \textit{w} to the denominator of original softmax function, and the refinement loss (RFloss) can be formulated as:
\begin{equation}\label{eq1}
RFloss = -\sum_{j=1}^{C} y_{j} \log \left(\tilde{p}_{j}\right)
\end{equation}

\begin{equation}\label{eq2}
\tilde{p}_{j}=R_{-}{softmax}\left(z_{j}\right)=\frac{e^{z_{j}}}{\sum_{k=1}^{C} {w}_{k} e^{z_{k}}}
\end{equation}
and \textit{C} is the number of categories including an extra class for background, \textit{z} is the output of the network. In practice, the ground truth label $y_j$ uses one-hot representation, and we have $\sum_{j=1}^{C} y_{j} =1$. For a proposal \textit{r}, reweighting factor $w_k$ is computed by:
\begin{equation}\label{eq3}
{w}_{k}=1-G(r) \cdot F_{k} \cdot\left(1-y_{k}\right)
\end{equation}
In this expression, \textit{G(r)} outputs 1 when \textit{G(r)} is a foreground region proposal and 0 when it belongs to background. And $F_k$ represents the frequency of category \textit{k} in the dataset, which outputs 1 when category \textit{k} belongs to novel classes and 0 when category \textit{k} belongs to base classes. 

\begin{figure}[h]
	\vskip 0.1in
	\begin{center}
		\centerline{\includegraphics[width=0.5\textwidth]{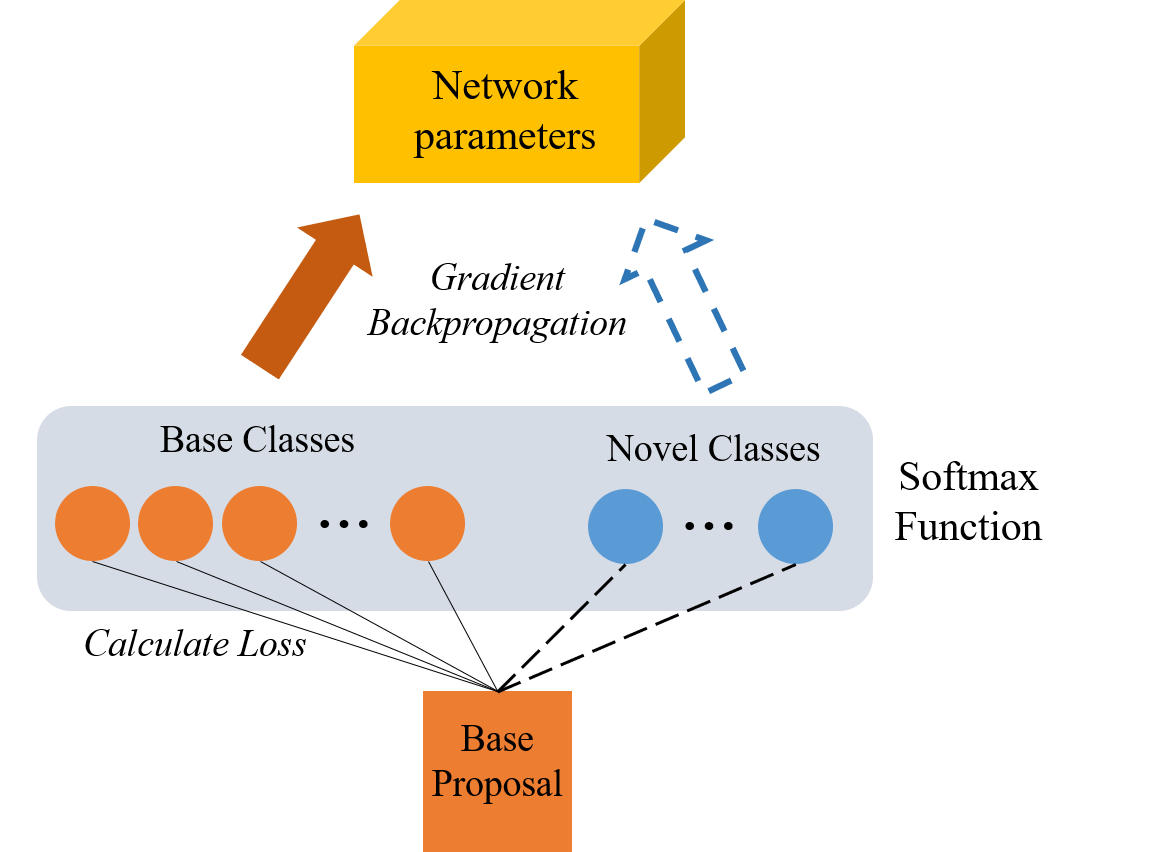}}
		\caption{The loss calculation of base proposal. The dotted line means novel classes do not participate in the loss calculation of base proposals and the dotted arrow indicates that the negative gradients would not be generated to hinder the learning of novel classes.}
		\label{fig3}
	\end{center}
	\vskip 0.1in
\end{figure}

As shown in Figure 3, if the proposal belongs to base classes, the computed loss for novel classes would be ignored during the gradient calculation. The generation mechanism of reweighting factor  $w_k$ is also shown in Algorithm 1.

\begin{algorithm}[ht]
	\caption{Generation Mechanism of Reweighting Factor  $w_k$ in Base Training Phase}
	\label{alg:1}
	\begin{algorithmic}
		\STATE {\bfseries Input:} Proposal $r$, Ground Truth Label Set \{$y_1$,$y_2$,..., $y_K$\}, and $K$ is the number of categories.
		\STATE {\bfseries Output:} Reweighting Factor Set \{$w_1$,$w_2$,..., $w_K$\}
		\FOR {iteration=$1$,..., $K$} 
		\IF{Proposal $r$ belongs to background}
		\STATE{$G(r)$ = 0, so $w_k$ = $1-0\cdot{F_k}\cdot(1-{y_k}) = 1$. When $w_k$ = 1, the mechanism of Refinement Loss is the same as the Softmax function.}
		\ELSE
		\IF{Proposal $r$ belongs to novel classes}
		\STATE{When the groundtruth label ${y_k} = 1$, the corresponding class belongs to novel classes, so ${F_k} = 1$ and $w_k = 1-1\cdot1\cdot(1-1) = 1$; When the groundtruth label ${y_k} = 0$, the corresponding class belongs to base classes, so ${F_k} = 0$ and $w_k = 1-1\cdot0\cdot(1-0) = 1$.}
		\ELSE
		\STATE{When the groundtruth label ${y_k} = 1$, the corresponding class belongs to base classes, so ${F_k} = 0$ and $w_k = 1-1\cdot0\cdot(1-1) = 1$; When the groundtruth label ${y_k} = 0$, the corresponding class belongs to novel classes, so ${F_k} = 1$ and $w_k = 1-1\cdot0\cdot(1-0) = 0$.}
		\ENDIF \ENDIF
		\ENDFOR
		\STATE \textbf{Return} \{$w_1$,$w_2$,..., $w_K$\}
	\end{algorithmic}
\end{algorithm}

The refinement loss could ease the negative impact of massive discouraging gradient on novel classes during the updating of parameters. Therefore, the model would be more sensitive to features of novel classes. And the base training model trained with refinement loss can also alleviate the misclassification of novel proposals in the beginning iterations of fine-tuning phase. We also provide various ablation studies and visualizations in Section 4.3. 

\subsection{Refinement Branch of RPN} Generating region proposals accounts for a large portion of the overall inference time, such as sliding window algorithm used in classical detection methods and SS (Selective Search) algorithm used in R-CNN\cite{RCNN}. In order to improve the proposal generation mechanism, Ren et al.\cite{Faster17} proposed to use region proposal network (RPN) to generate region proposals, which could not only shorten the overall inference time, but also improve the precision of the detection algorithm. However, the quantity and quality of region proposals could directly affect the detection precision of the corresponding category. In few-shot object detection task, the unbalanced distribution of region proposals is an important reason for the difficulty of improving the precision of novel classes. We modify the architecture of RPN, and attempt to alleviate the unbalanced proposal distribution. 

There are two prediction branches in the original RPN. Objectness branch obtains the \textit{objectness logits} of anchors through binary softmax cross entropy loss which is used to determine whether the anchor is positive or negative, and the regression branch is used to regress the relative offset of anchors to the ground truth boxes. The Find Top Proposal layer selects the first $K_1$ positive proposals according to the \textit{objectness logits} from high to low, and then removes duplicate and out-of-boundary proposals through NMS\cite{NMS} operations. Finally, only $K_2$ proposals are retained for classification and regression in the RoI head. We follow the original Faster R-CNN and set $K_1$=2000 and $K_2$=1000.

As for the few-shot object detection task, more novel proposals generated by RPN could improve the detection precision of novel classes. Based on this consideration, we propose the refinement branch for RPN as shown in Figure 4. The refinement branch optimizes \textit{bn (base-novel) logits} which could tell whether a proposal belongs to base classes or novel classes through binary softmax cross entropy loss. The way we generate the ground truth information of refinement branch is similar to that of objectness branch.

\begin{figure*}[t]
	\vskip 0.1in
	\begin{center}
		\centerline{\includegraphics[width=0.8\textwidth]{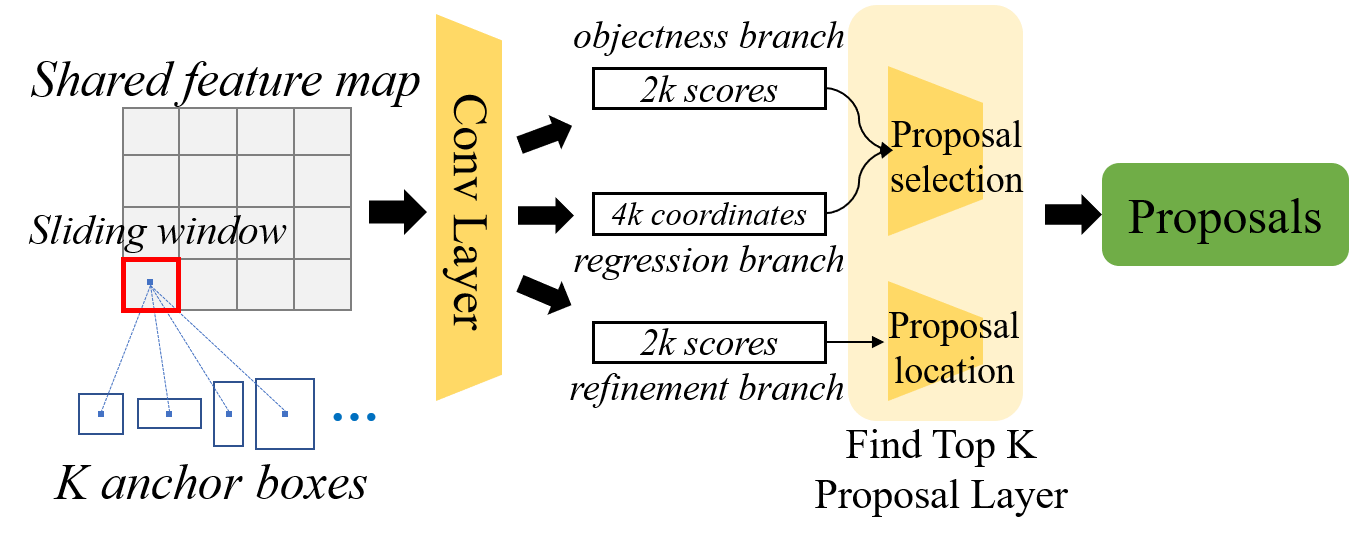}}
		\caption{The modified region proposal network (RPN).}
		\label{fig4}
	\end{center}
	\vskip -0.3in
\end{figure*}

For the ground truth information of refinement branch, a binary class label (of being a novel anchor or not) is assigned to each anchor. We assign a positive label to anchors that have an IoU (Intersection-over-Union) overlap higher than 0.7 with ground-truth boxes belong to novel classes. And an anchor would be assigned a negative label if its IoU ratio is higher than 0.7 with ground-truth box belong to base classes. An anchor which is neither positive nor negative would be assigned an ignore label and does not contribute to the gradient descent. With these definitions, we minimize a new loss function for RPN following the multi-task loss in Faster R-CNN. Our loss function for a sample is defined as:

\begin{large}
\begin{equation}\label{eq4}
\begin{array}{c}
L(\{ {p_{ob{j_i}}}\} ,\{ {t_i}\} ,\{ {p_{b{n_i}}}\} ) = \\
\frac{1}{{{N_{pos}}}}\sum\limits_i {{L_{cls}}({p_{ob{j_i}}},{p_{ob{j_i}}}^*)} \\
+ \alpha \frac{1}{{{N_{reg}}}}\sum\limits_i {{p_{ob{j_i}}}^*{L_{reg}}({t_i},{t_i}^*)} \\
+ \beta \frac{1}{{{M_{pos}}}}\sum\limits_i {{L_{ref}}(p\_b{n_i},p\_b{n_i}^*)} 
\end{array}
\end{equation}
\end{large}

The first two items of the loss function are exactly the same as the original RPN loss function. And the third item is specifically designed to optimize the refinement branch. Here, \textit{i} is the index of an anchor in a mini-batch. $p_{-}obj_{i}$ is the predicted probability of anchor i being an object and the ground-truth label $p_{-}obj_{i}^{*}$ is 1 if the anchor is positive, and is 0 if the anchor is negative. $t_{i}$ is a vector representing the 4 parameterized coordinates
of the predicted bounding box, and $t_{i}^{*}$ is that of the ground-truth box associated with a positive anchor. $p_{-}bn_{i}$ is the predicted probability of anchor \textit{i} being a novel anchor. The ground-truth label $p_{-}bn_{i}^{*}$ is 1 if the anchor is positive, and is 0 if the anchor is negative, and we would set the ground-truth label of the anchor to -1 if its IoU ratio is lower than 0.7, which represents an ignore label. The classification loss $L_{ref}$ is binary cross entropy loss over two classes (novel vs. base). The normalization term $M_{pos}$ is the number of selected anchor samples for refinement. By default, we set $\alpha=\beta=10$, and thus \textit{cls}, \textit{reg} and \textit{ref} terms are roughly equally weighted.

The \textit{bn logits} predicted by the refinement branch are used to help the selection of proposals. With the learning of the refinement branch, more proposals belonging to the novel classes will be generated during training. Formally, we define \textit{mixed logits} by the following formula:

\begin{equation}\label{eq5}
{mixed_{-}logits}= { objectness_{-}logits }+\theta *  { bn_{-}logits }
\end{equation}

A hyperparameter $\theta$ called mix factor is introduced to control the influence of \textit{bn logits} on the proposal screening.  $\theta$ is set to 1 over all our experiments and the different setting of  $\theta$ is studied in Section 4.3. The Find Top K Proposal Layer would select the first $K_1$ positive proposals according to the \textit{mixed logits} rather than the \textit{objectness logits}. The higher \textit{bn logits} scores, the proposal is more likely belong to the novel classes. And the higher \textit{mixed logits} score, the proposal is more likely be chosen for the final prediction. Therefore, \textit{mixed logits} could ensure more proposals belonging to the novel classes generated during the training of RPN, which could improve the detection accuracy of the novel classes.

\section{Experiments}
In this section, we present the simulation performance of our model. Following the setting of Kang et al.\cite{FSRW} and Wang et al.\cite{fsdet}. Our model is trained and evaluated on VOC\cite{VOC,VOC2012} and MS COCO\cite{COCO}. For VOC, our model is trained on the union set of VOC 2007 train\&val sets and VOC 2012 train\&val sets, and is evaluated on VOC 2007 test set. For COCO, our model is trained on COCO 2017 train set and evaluated on COCO 2017 val set. Through sufficient experiments, we have proved that the refinement of proposals during the training could effectively improve the AP(Average Precision) of novel classes. 

\subsection{Datasets and Basic Setup}
\textbf{Datasets Information.} For VOC, we select 5 classes as novel classes and make the remaining 15 classes as base classes. Our model is evaluated on 3 different base/novel splits. For fair comparison with baseline method, we set \{ bird, bus, cow, motorbike, sofa \} as novel classes in split set 1, \{ aero, bottle, cow, horse, sofa \} as novel classes in split set 2 and \{ boat, cat, motorbike, sheep, sofa \} as novel classes in split set 3. For COCO, the 20 classes overlapped with VOC is selected as novel classes form all 80 object classes, and the remaining 60 classes is set as the base classes. 

\begin{table*}[htb]
	\caption{Few-shot detection performance (mAP50) on the PASCAL VOC dataset. FRCN stands for Faster R-CNN. FsDet w/cos is FsDet with a cosine similarity based box classifier.}
	\vskip 0.1in
	\centering
	\begin{small}
		\renewcommand\arraystretch{1.5}
		\setlength{\tabcolsep}{1.7mm}{
			\begin{tabular}{c|ccccc|ccccc|ccccc}
				\toprule
				\multirow{2}{*}{Method / Shot} & \multicolumn{5}{c|}{Split Set 1} & \multicolumn{5}{c|}{Split Set 2} & \multicolumn{5}{c}{Split Set 3} \\  
				& 1    & 2    & 3    & 5    & 10   & 1    & 2    & 3    & 5    & 10   & 1    & 2    & 3    & 5    & 10   \\ \midrule
				YOLO-joint \cite{FSRW}                     & 0    & -    & 0    & 1.8  & 1.8  & 0    & -    & 0    & 1.8  & 0    & 1.8  & -    & 1.8  & 3.6  & 3.9  \\
				YOLO-ft \cite{FSRW}                        & 3.2  & -    & 6.4  & 7.5  & 12.3 & 8.2  & -    & 3.5  & 3.5  & 7.8  & 8.1  & -    & 7.6  & 9.5  & 10.5 \\
				YOLO-ft-full \cite{FSRW}                   & 6.6  & -    & 12.5 & 24.8 & 38.6 & 12.5 & -    & 11.6 & 16.1 & 33.9 & 13.0 & -    & 15.0 & 32.2 & 38.4 \\
				FSRW \cite{FSRW}                           & 14.8 & 15.5 & 26.7 & 33.9 & 47.2 & 15.7 & 15.3 & 22.7 & 30.1 & 40.5 & 21.3 & 25.6 & 28.4 & 42.8 & 45.9 \\ \midrule
				FRCN-joint \cite{MetaDet}                     & 0.3  & 0.0  & 1.2  & 0.9  & 1.7  & 0.0  & 0.0  & 1.1  & 1.9  & 1.7  & 0.2  & 0.5  & 1.2  & 1.9  & 2.8  \\
				FRCN-joint-ft \cite{MetaDet}                 & 9.1  & 10.9 & 13.7 & 25.0 & 39.5 & 10.9 & 13.2 & 17.6 & 19.5 & 36.5 & 15.0 & 15.1 & 18.3 & 33.1 & 35.9 \\
				MetaDet \cite{MetaDet}                        & 18.9 & 20.6 & 30.2 & 36.8 & 49.6 & 21.8 & 23.1 & 27.8 & 31.7 & 43.0 & 20.6 & 23.9 & 29.4 & 43.9 & 44.1 \\ \midrule
				FRCN+joint \cite{MetaRCNN}                     & 2.7  & 3.1  & 4.3  & 11.8 & 29.0 & 1.9  & 2.6  & 8.1  & 9.9  & 12.6 & 5.2  & 7.5  & 6.4  & 6.4  & 6.4  \\
				FRCN+ft \cite{MetaRCNN}                       & 11.9 & 16.4 & 29.0 & 36.9 & 36.9 & 5.9  & 8.5  & 23.4 & 29.1 & 28.8 & 5.0  & 9.6  & 18.1 & 30.8 & 43.4 \\
				FRCN+ft-full \cite{MetaRCNN}                  & 13.8 & 19.6 & 32.8 & 41.5 & 45.6 & 7.9  & 15.3 & 26.2 & 31.6 & 39.1 & 9.8  & 11.3 & 19.1 & 35.0 & 45.1 \\
				Meta R-CNN \cite{MetaRCNN}                    & 19.9 & 25.5 & 35.0 & 45.7 & 51.5 & 10.4 & 19.4 & 29.6 & 34.8 & 45.4 & 14.3 & 18.2 & 27.5 & 41.2 & 48.1 \\ \midrule
				FsDet w/cos \cite{fsdet}                    & 39.8 & 36.1 & 44.7 & 55.7 & 56.0 & 23.5 & 26.9 & 34.1 & 35.1 & 39.1 & 30.8 & 34.8 & 42.8 & 49.5 & 49.8 \\
				FsDet w/cos(Our Impl.)                      & 26.3 & 40.6 & 44.9 & 49.9 & 52.9 & 23.4 & 26.6 & 33.4 & 32.7 & 40.1 & 15.5 & 28.8 & 33.0 & 46.9 & 45.2 \\ \midrule
				\begin{tabular}[c]{@{}c@{}}Ours\\ \textit{Compare with Our Impl.}\end{tabular}                           & \begin{tabular}[c]{@{}c@{}}27.9\\ +$1.6$\end{tabular} & \begin{tabular}[c]{@{}c@{}}\textbf{41.8}\\ +$1.2$\end{tabular} & \begin{tabular}[c]{@{}c@{}}\textbf{48.3}\\ +$3.4$\end{tabular} & \begin{tabular}[c]{@{}c@{}}55.2\\ +$5.3$\end{tabular} & \begin{tabular}[c]{@{}c@{}}\textbf{56.2}\\ +$3.3$\end{tabular} & \begin{tabular}[c]{@{}c@{}}\textbf{24.5}\\ +$1.1$\end{tabular} & \begin{tabular}[c]{@{}c@{}}\textbf{28.3}\\ +$1.7$\end{tabular} & \begin{tabular}[c]{@{}c@{}}\textbf{36.9}\\ +$3.5$\end{tabular} & \begin{tabular}[c]{@{}c@{}}\textbf{36.2}\\ +$3.5$\end{tabular} & \begin{tabular}[c]{@{}c@{}}\textbf{45.0}\\ +$4.9$\end{tabular}  & \begin{tabular}[c]{@{}c@{}}15.6\\ +$0.1$\end{tabular}  & \begin{tabular}[c]{@{}c@{}}30.2\\ +$1.4$\end{tabular}  & \begin{tabular}[c]{@{}c@{}}34.1\\ +$1.1$\end{tabular}  & \begin{tabular}[c]{@{}c@{}}48.7\\ +$1.8$\end{tabular}  & \begin{tabular}[c]{@{}c@{}}\textbf{50.3}\\ +$5.1$\end{tabular}  \\ \bottomrule
		\end{tabular}}
	\end{small}
	\vskip 0.1in
\end{table*}

\textbf{Implementation Details.} All experiments are conducted with CUDA 10.0, cuDNN v7.4 and PyTorch 1.3.0. Our models are trained using SGD with momentum 0.9. The batch size of base training phase and the fine-tuning phase is set to be 4 and 16 on NVIDIA GTX 1080Ti. We use Faster R-CNN\cite{Faster17} as our base detector and Resnet-101\cite{ResNet101} with a Feature Pyramid Network\cite{FPN} as the feature extractor. A learning rate of 0.02 is used during base training phase and 0.001 during few-shot fine-tuning phase. As for other training parameters, we follow the few-shot object detection settings introduced in FsDet.

\begin{table*}[htb]	
	\caption{Few-shot detection performance for the novel categories on the COCO dataset.}
	\vskip 0.1in
	\centering
	\begin{small}
		\renewcommand\arraystretch{1.5}
		\begin{tabular}{cc|ccc|ccc}
			\toprule
			&                          & \multicolumn{6}{c}{Average Precision}     \\ \midrule
			\multicolumn{1}{c|}{Shots}               & Method                   & 0.5:0.95 & 0.5  & 0.75 & S   & M    & L    \\ \midrule
			\multicolumn{1}{c|}{\multirow{7}{*}{10}} & FSRW \cite{FSRW}            & 5.6      & 12.3 & 4.6  & 0.9 & 3.5  & 10.5 \\
			\multicolumn{1}{c|}{}                    & MetaDet \cite{MetaDet}          & 7.1      & 14.6 & 6.1  & 1.0 & 4.1  & 12.2 \\
			\multicolumn{1}{c|}{}                    & FRCN+ft-full \cite{MetaRCNN}    & 6.5      & 13.4 & 5.9  & 1.8 & 5.3  & 11.3 \\
			\multicolumn{1}{c|}{}                    & Meta R-CNN \cite{MetaRCNN}              & 8.7      & 19.1 & 6.6  & 2.3 & 7.7  & 14.0 \\
			\multicolumn{1}{c|}{}                    & FsDet w/cos \cite{fsdet}              & 10.0     & -    & 9.2  & -   & -    & -    \\
			\multicolumn{1}{c|}{}                    & FsDet w/cos  (Our Impl.) & 9.1          & 16.5     & 9.2     & 3.4    & 7.7     & 13.5     \\
			\multicolumn{1}{c|}{}                    & \begin{tabular}[c]{@{}c@{}}Ours\\ \textit{Compare with Our Impl.}\end{tabular}                     & \begin{tabular}[c]{@{}c@{}}9.4\\ +$0.3$\end{tabular}          & \begin{tabular}[c]{@{}c@{}}17.4\\ +$0.9$\end{tabular}     & \begin{tabular}[c]{@{}c@{}}\textbf{9.3}\\ +$0.1$\end{tabular}     & \begin{tabular}[c]{@{}c@{}}3.3\\ -$0.1$\end{tabular}    & \begin{tabular}[c]{@{}c@{}}\textbf{8.0}\\ +$0.3$\end{tabular}     & \begin{tabular}[c]{@{}c@{}}\textbf{14.2}\\ +$0.7$\end{tabular}     \\ \midrule
			\multicolumn{1}{c|}{\multirow{7}{*}{30}} & FSRW \cite{FSRW}            & 9.1      & 19.0 & 7.6  & 0.4 & 2.9  & 12.3 \\
			\multicolumn{1}{c|}{}                    & MetaDet \cite{MetaDet}          & 11.3     & 21.7 & 8.1  & 1.1 & 6.2  & 17.3 \\
			\multicolumn{1}{c|}{}                    & FRCN+ft-full \cite{MetaRCNN}    & 11.1     & 21.6 & 10.3 & 2.9 & 8.8  & 18.9 \\
			\multicolumn{1}{c|}{}                    & Meta R-CNN \cite{MetaRCNN}              & 12.4     & 25.3 & 10.8 & 2.8 & 11.6 & 19.0 \\
			\multicolumn{1}{c|}{}                    & FsDet w/cos \cite{fsdet}             & 13.7     & -    & 13.4 & -   & -    & -    \\
			\multicolumn{1}{c|}{}                    & FsDet w/cos  (Our Impl.) & 12.8         & 23.8     & 12.9      & 4.8    & 10.8     & 19.1     \\
			\multicolumn{1}{c|}{}                    & \begin{tabular}[c]{@{}c@{}}Ours\\ \textit{Compare with Our Impl.}\end{tabular}                     & \begin{tabular}[c]{@{}c@{}}13.6\\ +$0.8$\end{tabular}         & \begin{tabular}[c]{@{}c@{}}\textbf{25.6}\\ +$1.8$\end{tabular}     & \begin{tabular}[c]{@{}c@{}}\textbf{13.5}\\ +$0.6$\end{tabular}     & \begin{tabular}[c]{@{}c@{}}\textbf{4.8}\\ +$0.0$\end{tabular}    & \begin{tabular}[c]{@{}c@{}}\textbf{11.9}\\ +$1.1$\end{tabular}     & \begin{tabular}[c]{@{}c@{}}\textbf{20.9}\\ +$1.8$\end{tabular}     \\ \bottomrule
		\end{tabular}
	\end{small}
	\vskip 0.1in
\end{table*}

\textbf{Baselines.} Meta learning, metric-learning and transfer learning have provided different directions for improvement on few-shot object detection algorithms. There are several object detection algorithms have already achieved considerable performance under few-shot scenarios, such as FSRW\cite{FSRW}, Meta-RCNN\cite{MetaRCNN} and RepMet\cite{RepMet}. However, FsDet achieves the state-of-the-art with an intuitive and simple training strategy based on transfer learning. FsDet adopts a two-phase training scheme, which consists of base training phase and fine-tuning phase. In the base training phase, the entire object detector is trained on the data abundant base classes, and in the fine-tuning phase only the last classification layer and regression layer of the detector are trained on a small balanced training set consisting of both base and novel classes while freezing the other parameters of the model including RPN. So, we choose FsDet as the baseline method and conduct further research on proposal generation during training. 

Due to limited computing resources, we set the batch size as 4 in the base training phase. However, the batch size of FsDet in the base training phase is 16 and they train the model on 8 GPUs. Under few-shot setting, the samples of novel classes and base classes in fine-tuning phase is selected from the official datasets by different random seed. And different random seed would select different training samples, the feature scale and quality of those samples may vary to a certain extent and cause the detection results to fluctuate. Due to the different experimental settings between the baselines method and our method, for fair comparison, we retrain the model of FsDet based on the official released code under our experimental environment. The result of retrained model in result tables will be followed by the phrase "Our Impl.".

\subsection{Comparison with Baselines}
We compare our results with the baseline method and other SOTA few-shot detection methods based on meta learning and metric-learning. 

\textbf{Results on PASCAL VOC.} We present our main results on novel classes with three different splits in Table 1. It can be seen from this table that the results outperform previous approaches under different few-shot settings. When the number of instances gradually increases, our method could bring stability and solid improvement on accuracy. Specifically, by solving unbalanced distribution of proposals, we surpass the baseline method implemented on our machine by roughly 1\%$\sim$6\% on PASCAL VOC benchmark. It clearly highlights the effectiveness of proposal refinement approach.

\textbf{Results on COCO.} Table 2 shows that the AP of our method on novel classes on 10-shot setting and 30-shot setting. Although COCO is quite challenging, we still achieve an increase of 0.8\% on 30-shot compared with FsDet. Our method generates more novel proposals, which could help the network locate more novel instances, so we have a big improvement on $nAP_{50}$. 

\textbf{Refinement of proposals.} In order to improve the detection accuracy of FSOD (few-shot object detector), we focus on the generation mechanism of proposals.  We propose refinement loss in base training phase and refinement branch in fine-tuning phase to alleviate the imbalance between novel proposals and base proposals. Figure 5 illustrates an overview of misclassification of foreground proposals belonging to the novel classes in the fine-tuning phase. 
\begin{figure}[h]
	\vskip -0.1in
	\begin{center}
		\centerline{\includegraphics[width=0.5\textwidth]{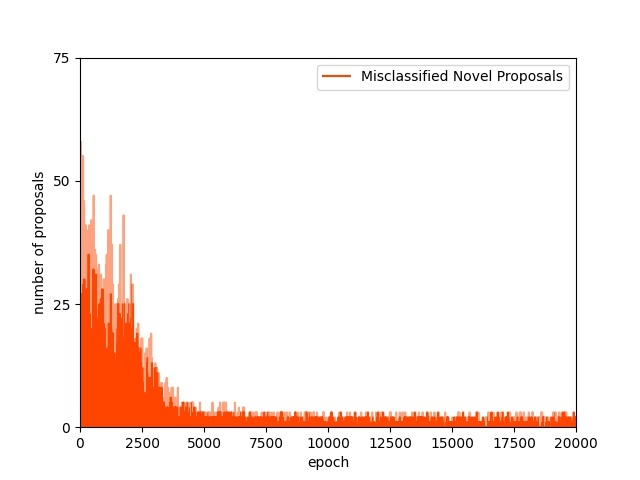}}
		\caption{The visualization of novel proposals misclassified by baseline method and our method under 5-shot setting. The light-colored orange bar represents number of misclassified novel proposals of the baseline method and the deep-colored orange bar represents for our proposed method.}
		\label{fig5}
	\end{center}
	\vskip 0.1in
\end{figure}

\begin{figure*}[h]
	\begin{center}
		\centerline{\includegraphics[width=1\textwidth]{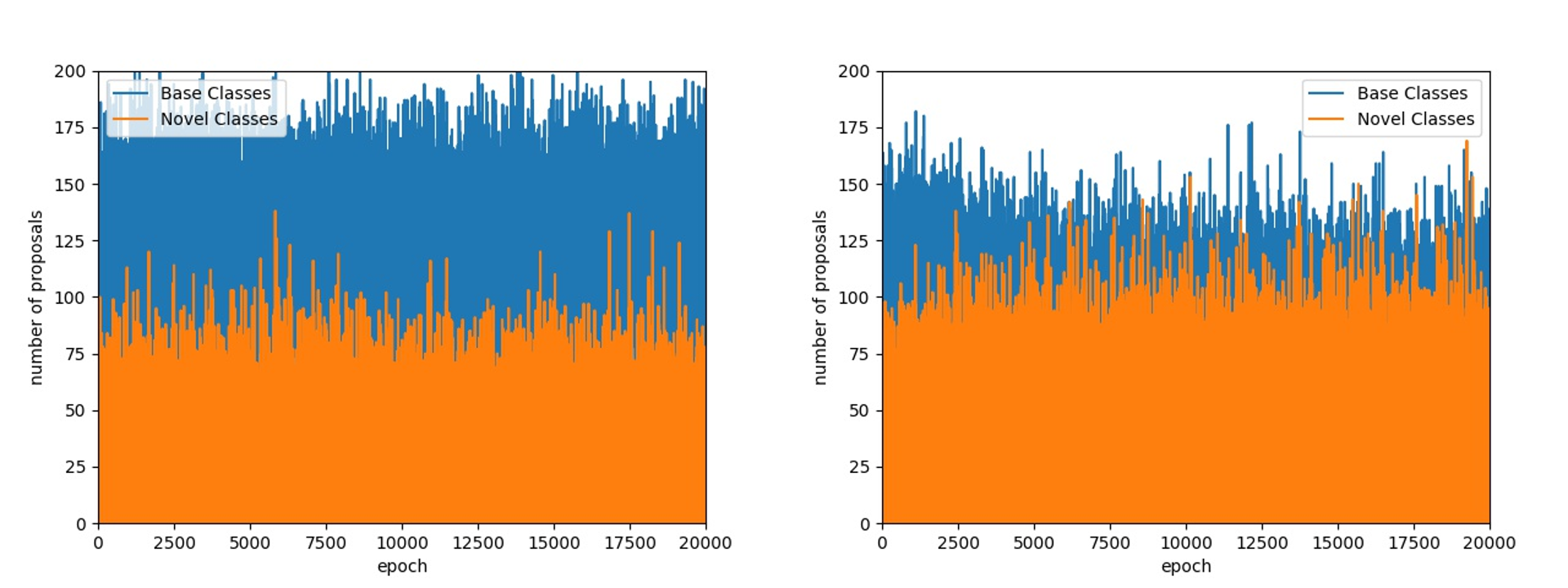}}
		\caption{The number of foreground proposals of novel classes and base classes during the fine-tuning phase under 5-shot setting. The figure on the left represents the baseline method and the figure on the right represents our proposed method.}
		\label{fig6}
	\end{center}
	\vskip 0.1in
\end{figure*}

It is worth noting that in the first 2000 epoch of baseline method, a large proportion of the proposals belonging to novel classes were incorrectly classified as base classes. The main reason of the misclassification is that the base training model used for finetune is only trained on samples of base classes, thus the base training model is insensitive to the features of novel classes. Since the model is more sensitive to base classes, the prediction score of novel proposals will be suppressed by base proposals after being processed by softmax function in the beginning of fine-tuning phase. The suppression of base classes on novel classes is also emphasized by Li et al.\cite{group_softmax}. In our proposed method, the misclassification in the beginning of the fine-tuning phase has been alleviated due to the application of refinement loss during the base training phase.

Figure 6 gives a comparison of the number of foreground proposals between base classes and novel classes during the fine-tuning phase. It can be seen that in the baseline method, the number of base proposals have an overwhelming advantage during the entire training period which is also an important cause of the difficulty of improving AP of novel classes. In our proposed method, the dominant advantage of the base proposal has been greatly weakened and the number of proposal changes dynamically with the training of the network. As the training progresses, the number of novel proposals gradually increases, and the number of base proposals gradually decreases, which accounts for the increasing AP of novel class. 

\subsection{Ablation Studies}

We analyze the effect of various modules in our method by comparing the performance on both base classes and novel classes. The experiments are conducted on PASCAL VOC split set 1 if not specified.

\textbf{Effectiveness of refinement loss.} We evaluate the contribution of the proposed refinement loss under different few-shot setting for better reference. Results are reported in Table 3, where RFloss denotes the refinement loss.

\begin{table}[h]
	\caption{Ablation of Refinement Loss of the box classifier.}
	\vskip 0.1in
	\centering
	\begin{small}
		\begin{tabular}{ccc|cccc}
			\toprule
			Shots               & Baseline & RFloss & bAP  & $bAP_{50}$ & nAP  & $nAP_{50}$ \\ \midrule
			\multirow{2}{*}{1}  & \Checkmark       &        & 50.3 & 77.6  & 16.1 & 26.3  \\
			& \Checkmark        & \Checkmark      & 50.1 & 76.2  & \textbf{16.5} & \textbf{27.1}  \\ \midrule
			\multirow{2}{*}{5}  & \Checkmark        &        & 50.5 & 77.4  & 29.7 & 49.9  \\
			& \Checkmark        & \Checkmark      & 48.4 & 74.9  & \textbf{31.3} & \textbf{51.6}  \\ \midrule
			\multirow{2}{*}{10} & \Checkmark        &        & 50.9 & 78.4  & 32.1 & 52.9  \\
			& \Checkmark        & \Checkmark      & 47.6 & 75.7  & \textbf{33.6} & \textbf{55.4}  \\ \bottomrule
		\end{tabular}
	\end{small}
	\vskip 0.1in
\end{table}

From Table 3, it can be seen that the refinement loss contribute to accuracy gain of novel class under different setting. But the use of RFloss led to slight decline of bAP, since the RFloss would interface the gradients backpropagation of base classes. However, in the few-shot object detection task, we always pay more attention to the accuracy of novel classes, so it is acceptable to use the reduction of bAP in exchange for the improvement of nAP, especially in real application scenarios.

\textbf{Freeze different module during fine-tuning phase.} We analyzed the accuracy changes brought by freezing different modules during the fine-tuning phase and results is given in Table 4. For fair comparison, the base training model is obtained by the training strategy of the baseline method, and fine-tuning on the PASCAL VOC split set 1, with 5-shot setting.

\begin{table}[h]
	\caption{Ablation of different training strategy. FZR means that the module is frozen, the check mark means that the module participated in the parameter during training and backbone is frozen by default.}
	\vskip 0.1in
	\centering
	\begin{small}
		\begin{tabular}{ccc|cc}
			\toprule
			RPN & \begin{tabular}[c]{@{}c@{}}RoI Feature\\ Extractor\end{tabular} & \begin{tabular}[c]{@{}c@{}}Box Classifier +\\  Box  Regressor\end{tabular} & nAP  & nAP50 \\ \midrule
			FRZ & FRZ                                                               & \Checkmark                                                                              & 29.7 & 49.9  \\
			FRZ & \Checkmark                                                                 & \Checkmark                                                                              & 26.6 & 42.9  \\
			\Checkmark   & \Checkmark                                                                 & \Checkmark                                                                              & 26.5 & 44.2  \\
			\Checkmark   & FRZ                                                               & \Checkmark                                                                              & \textbf{29.9} & \textbf{51.0}  \\ \bottomrule
		\end{tabular}
	\end{small}
	\vskip 0.1in
\end{table}

The first row in Table 4 represents the training strategy of the baseline method in the fine-tuning phase, which is, freezing all modules except the classification and regression layers. The second row shows that unfreeze the RoI Feature Extractor would bring decrease on the accuracy of novel classes. It is worth noting that $nAP_{50}$ increased by 1.1\% after unfreezing the RPN compare to the baseline, which indicates that updating parameters of RPN in the fine-tuning phase might generate more novel proposals. Since it has optimal performance on novel classes, we unfreeze RPN in all of our experiments.

\textbf{Mix factor of Refinement Branch.} We explore the effect of different mix factors for computing mixed logits. We compare four different factors, $\theta$ = 0; 0.1; 1; 10. We use the same evaluation setting as the previous ablation study and report the results in Table 5.

\begin{table}[h]
	\caption{Ablation of different mix factor of Refinement Branch.}
	\vskip 0.1in
	\centering
	\begin{small}
		\begin{tabular}{c|cccccc}
			\toprule
			Mix Factor & bAP  & $bAP_{50}$ & nAP  & $nAP_{50}$ \\ \midrule
			0          & \textbf{50.5} & \textbf{77.4}   & 29.7 & 49.9   \\
			0.1        & 49.1 & 76.8   & 29.6 & 49.3   \\
			1          & 32.5 & 59.7  & \textbf{30.6} & \textbf{52.5}   \\
			10         & 20.3 & 34.1   & 25.9 & 37.4  \\ \bottomrule
		\end{tabular}
	\end{small}
	\vskip 0.1in
\end{table}

\begin{figure}[h]
	\vskip -0.1in
	\begin{center}
		\centerline{\includegraphics[width=0.5\textwidth]{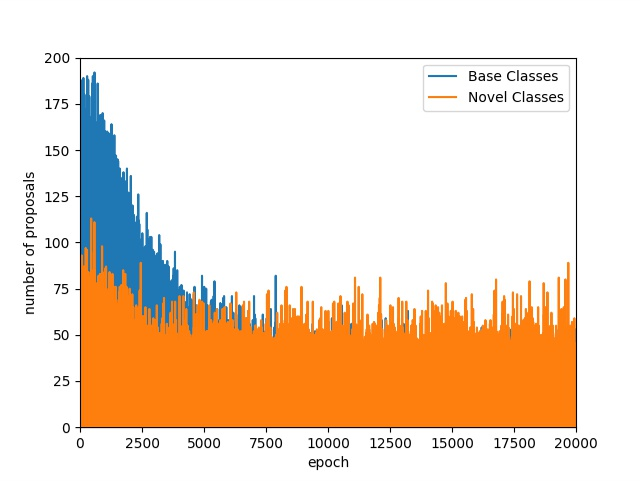}}
		\caption{The number of foreground proposals of novel classes and base classes $\theta$ = 10 during the finetune phase under 5-shot setting.}
		\label{fig7}
	\end{center}
	\vskip 0.2in
\end{figure}

It can be seen that when $\theta$ = 0.1, nAP is hardly affected by \textit{bn logits}. When $\theta$ = 10, the influence of \textit{bn logits} on \textit{mixed logits} dominates, resulting in a significant decrease in the accuracy of bAP. However, more proposals belonging to novel classes didn’t significantly increase nAP, since there are only a few novel instances under few-shot setting, when the novel proposals generated by the RPN dominates, the reduction of nAP caused by overfitting is easy to occur. $\theta$ = 1 outperforms the other mix factors in novel AP, so we use this setting in all of our experiments related with refinement branch. As shown in Figure 7, the number of base proposals is severely suppressed with the learning of refinement branch when  $\theta$ = 10, which also proves the effectiveness of the proposed refinement branch.

\section{Conclusion}
In this paper, we discuss the influence of the unbalanced proposal distribution between novel classes and base classes in few-shot object detection task. We propose the refinement branch for RPN to rebalance the proposal distribution. In addition, the refinement loss is introduced to obtain a stronger base training model for novel classes. Simulation results show that our method has the ability to improve the detection precision of novel classes with negligible increase in complexity and the state-of-the-art performance demonstrates the superiority of our proposal refinement approach.


%

\section*{Acknowledgment}
This work has been supported by the National Natural Science Foundation of China (Nos.61772387, 62071354), the National Natural Science Foundation of Shaanxi Province (Grant No.2019ZDLGY03-03), and also supported by the ISN State Key Laboratory.

\ifCLASSOPTIONcaptionsoff
  \newpage
\fi



\bibliographystyle{IEEEtran}
\bibliography{bare_jrnl}

@ARTICLE{Faster17,
  author={S. {Ren} and K. {He} and R. {Girshick} and J. {Sun}},
  journal={In Advances in IEEE Transactions on Pattern Analysis and Machine Intelligence}, 
  title={Faster R-CNN: Towards Real-Time Object Detection with Region Proposal Networks}, 
  year={2017},
  volume={39},
  number={6},
  doi={10.1109/TPAMI.2016.2577031},
  pages={1137-1149}}

@INPROCEEDINGS{yolov217,
  author={J. {Redmon} and A. {Farhadi}},
  booktitle={Proceedings of the IEEE Conference on Computer Vision and Pattern Recognition}, 
  title={YOLO9000: Better, Faster, Stronger}, 
  year={2017},
  volume={},
  number={},
  doi={10.1109/CVPR.2017.690},
  pages={6517-6525}}

@inproceedings{centernet,
  author={ Zhou, Xingyi  and  Wang, Dequan  and Krhenbühl, Philipp},
  booktitle={Proceedings of the IEEE Conference on Computer Vision and Pattern Recognition}, 
  title={Objects as points}, 
  year={2019},
  volume={},
  number={},
  doi={},
  pages={4367-4375}}

@inproceedings{MAML,
author = {Finn, Chelsea and Abbeel, Pieter and Levine, Sergey},
booktitle = {Proceedings of the 34th International Conference on Machine Learning - Volume 70},
title = {Model-Agnostic Meta-Learning for Fast Adaptation of Deep Networks},
year = {2017},
publisher = {JMLR.org},
  doi={10.5555/3305381.3305498},
pages = {1126–1135}}

@INPROCEEDINGS{L2C,
  author={F. {Sung} and Y. {Yang} and L. {Zhang} and T. {Xiang} and P. H. S. {Torr} and T. M. {Hospedales}},
  booktitle={Proceedings of the IEEE Conference on Computer Vision and Pattern Recognition}, 
  title={Learning to Compare: Relation Network for Few-Shot Learning}, 
  year={2018},
  volume={},
  number={},
  doi={10.1109/CVPR.2018.00131},
  pages={1199-1208}}

@inproceedings{matching,
author = {Vinyals, Oriol and Blundell, Charles and Lillicrap, Timothy and Kavukcuoglu, Koray and Wierstra, Daan},
booktitle = {Advances in Neural Information Processing Systems},
title = {Matching Networks for One Shot Learning},
year = {2016},
month = {},
  doi={},
pages = {3630-3638}}

@inproceedings{Prototypical,
author = {Snell, Jake and Swersky, Kevin and Zemel, Richard},
booktitle = {Advances in Neural Information Processing Systems},
title ={Prototypical Networks for Few-shot Learning},
year = {2017},
month = {},
  doi={},
pages = {4077-4087}}

@ARTICLE{D2N4,
  author={X. {Yang} and X. {Nan} and B. {Song}},
  journal={IEEE Transactions on Geoscience and Remote Sensing}, 
  title={D2N4: A Discriminative Deep Nearest Neighbor Neural Network for Few-Shot Space Target Recognition}, 
  year={2020},
  volume={58},
  number={5},
  doi={10.1109/TGRS.2019.2959838},
  pages={3667-3676}}

@INPROCEEDINGS{FSRW,
  author={B. {Kang} and Z. {Liu} and X. {Wang} and F. {Yu} and J. {Feng} and T. {Darrell}},
  booktitle={Proceedings of the IEEE International Conference on Computer Vision}, 
  title={Few-Shot Object Detection via Feature Reweighting}, 
  year={2019},
  volume={},
  number={},
  doi={10.1109/ICCV.2019.00851},
  pages={8419-8428}}

@INPROCEEDINGS{RepMet,
  author={L. {Karlinsky} and J. {Shtok} and S. {Harary} and E. {Schwartz} and A. {Aides} and R. {Feris} and R. {Giryes} and A. M. {Bronstein}},
  booktitle={Proceedings of the IEEE Conference on Computer Vision and Pattern Recognition}, 
  title={RepMet: Representative-Based Metric Learning for Classification and Few-Shot Object Detection}, 
  year={2019},
  volume={},
  number={},
  doi={10.1109/CVPR.2019.00534},
  pages={5192-5201}}

@ARTICLE{fsdet,
author = {Wang, Xin and Huang, Thomas and Darrell, Trevor and Gonzalez, Joseph and Yu, Fisher},
journal = {arXiv preprint arXiv:2003.06957},
title = {Frustratingly Simple Few-Shot Object Detection},
year = {2020},
month = {},
  doi={},
pages = {}}

@inproceedings{VOC,
  title={The PASCAL Visual Object Classes Challenge 2009 (VOC2009) Results},
  author={ Everingham, M. },
  booktitle={http://www.pascal-etwork.org/challenges/VOC/voc2009/workshop/index.html},
  doi={},
  year={2007}
}

@inproceedings{COCO,
  title={Microsoft COCO: Common Objects in Context},
  author={ Lin, Tsung Yi  and  Maire, Michael  and  Belongie, Serge  and  Hays, James  and  Zitnick, C. Lawrence },
  booktitle={ECCV},
  doi={10.1007/978-3-319-10602-1_48},
  year={2014},
}

@INPROCEEDINGS{group_softmax,
  author={Y. {Li} and T. {Wang} and B. {Kang} and S. {Tang} and C. {Wang} and J. {Li} and J. {Feng}},
  booktitle={Proceedings of the IEEE Conference on Computer Vision and Pattern Recognition}, 
  title={Overcoming Classifier Imbalance for Long-Tail Object Detection With Balanced Group Softmax}, 
  year={2020},
  volume={},
  doi={10.1109/CVPR42600.2020.01100},
  number={},
  pages={10988-10997}}

@ARTICLE{Decoupling,
  title={Decoupling Representation and Classifier for Long-Tailed Recognition},
  journal = {arXiv preprint arXiv:1910.09217},
  author={ Kang, Bingyi  and  Xie, Saining  and  Rohrbach, Marcus  and  Yan, Zhicheng  and  Gordo, Albert  and  Feng, Jiashi  and  Kalantidis, Yannis },
  pages = {},
  doi={},
  year={2019},
}

@ARTICLE{Self_Supervised,
author = {Chen, Da and Chen, Yuefeng and Li, Yuhong and Mao, Feng and He, Yuan and Xue, Hui},
journal = {arXiv preprint arXiv:1911.06045},
year = {2019},
month = {},
pages = {},
  doi={},
title = {Self-Supervised Learning For Few-Shot Image Classification}
}

@INPROCEEDINGS{FPN,
  author={T. {Lin} and P. {Dollár} and R. {Girshick} and K. {He} and B. {Hariharan} and S. {Belongie}},
  booktitle={Proceedings of the IEEE Conference on Computer Vision and Pattern Recognition}, 
  title={Feature Pyramid Networks for Object Detection}, 
  year={2017},
  volume={},
  number={},
  doi={10.1109/CVPR.2017.106},
  pages={936-944}}

@INPROCEEDINGS{Cascade,
  author={Z. {Cai} and N. {Vasconcelos}},
  booktitle={Proceedings of the IEEE Conference on Computer Vision and Pattern Recognition}, 
  title={Cascade R-CNN: Delving Into High Quality Object Detection}, 
  year={2018},
  volume={},
  number={},
  doi={10.1109/CVPR.2018.00644},
  pages={6154-6162}}

@ARTICLE{Mask,
  author={K. {He} and G. {Gkioxari} and P. {Dollár} and R. {Girshick}},
  journal={IEEE Transactions on Pattern Analysis and Machine Intelligence}, 
  title={Mask R-CNN}, 
  year={2020},
  volume={42},
  number={2},
  doi={10.1109/TPAMI.2018.2844175},
  pages={386-397}}

@INPROCEEDINGS{Fast,
  author={R. {Girshick}},
  booktitle={Proceedings of the IEEE International Conference on Computer Vision}, 
  title={Fast R-CNN}, 
  year={2015},
  volume={},
  number={},
  doi={10.1109/ICCV.2015.169},
  pages={1440-1448}}

@ARTICLE{CornerNet,
author = {Law, Hei and Deng, Jia},
journal={International Journal of Computer Vision},
year = {2020},
month = {},
pages = {642–656},
  doi={10.1007/978-3-030-01264-9_45},
title = {CornerNet: Detecting Objects as Paired Keypoints}}

@ARTICLE{focalloss,
  author={T. {Lin} and P. {Goyal} and R. {Girshick} and K. {He} and P. {Dollár}},
  journal={IEEE Transactions on Pattern Analysis and Machine Intelligence}, 
  title={Focal Loss for Dense Object Detection}, 
  year={2020},
  volume={42},
  number={2},
  doi={10.1109/ICCV.2017.324},
  pages={318-327}}

@INPROCEEDINGS{yolov1,
  author={J. {Redmon} and S. {Divvala} and R. {Girshick} and A. {Farhadi}},
  booktitle={Proceedings of the IEEE Conference on Computer Vision and Pattern Recognition}, 
  title={You Only Look Once: Unified, Real-Time Object Detection}, 
  year={2016},
  volume={},
  number={},
  doi={10.1109/CVPR.2016.91},
  pages={779-788}}

@article{YOLOv3,
  title={YOLOv3: An Incremental Improvement},
  author={ Redmon, Joseph  and  Farhadi, Ali },
  journal={arXiv preprint 	arXiv:1804.02767},
  doi={},
  year={2018}}

@INPROCEEDINGS{RCNN,
  author={R. {Girshick} and J. {Donahue} and T. {Darrell} and J. {Malik}},
  booktitle={Proceedings of the IEEE Conference on Computer Vision and Pattern Recognition}, 
  title={Rich Feature Hierarchies for Accurate Object Detection and Semantic Segmentation}, 
  year={2014},
  volume={},
  number={},
  doi={10.1109/CVPR.2014.81},
  pages={580-587}}

@ARTICLE{DAPNA,
author = { Guan, Jiechao  and  Lu, Zhiwu  and  Xiang, Tao  and  Wen, Ji Rong },
journal = {arXiv preprint arXiv:2002.02050},
title = {Few-Shot Learning as Domain Adaptation: Algorithm and Analysis},
year = {2020},
month = {},
  doi={},
pages = {}}

@INPROCEEDINGS{LSTD,
author = {Chen, Hao and Wang, Yali and Wang, Guoyou and Qiao, Yu},
booktitle = {Proceedings of the 32th AAAI Conference on Artificial Intelligence },
year = {2018},
month = {},
pages = {},
  doi={},
title = {LSTD: A Low-Shot Transfer Detector for Object Detection}
}

@INPROCEEDINGS {MetaRCNN,
author = {X. Yan and Z. Chen and A. Xu and X. Wang and X. Liang and L. Lin},
booktitle = {Proceedings of the IEEE International Conference on Computer Vision},
title = {Meta R-CNN: Towards General Solver for Instance-Level Low-Shot Learning},
year = {2019},
issn = {},
  doi={10.1109/ICCV.2019.00967},
pages = {9576-9585}}

@INPROCEEDINGS{EQloss,
  author={J. {Tan} and C. {Wang} and B. {Li} and Q. {Li} and W. {Ouyang} and C. {Yin} and J. {Yan}},
  booktitle={Proceedings of the IEEE Conference on Computer Vision and Pattern Recognition}, 
  title={Equalization Loss for Long-Tailed Object Recognition}, 
  year={2020},
  volume={},
  number={},
  doi={10.1109/CVPR42600.2020.01168},
  pages={11659-11668}}

@INPROCEEDINGS{NMS,
  author={A. {Neubeck} and L. {Van Gool}},
  booktitle={Proceedings of the 18th International Conference on Pattern Recognition}, 
  title={Efficient Non-Maximum Suppression}, 
  year={2006},
  volume={3},
  number={},
  doi={10.1109/ICPR.2006.479},
  pages={850-855}}

@article{VOC2012,
  title={The Pascal Visual Object Classes Challenge: A Retrospective},
  author={ Everingham, Mark  and  Eslami, S. M. Ali  and  Van Gool, Luc  and  Williams, Christopher K. I.  and  Winn, John  and  Zisserman, Andrew },
  journal={International Journal of Computer Vision},
  volume={111},
  number={1},
  pages={98-136},
  doi={10.1007/s11263-014-0733-5},
  year={2015},
}

@INPROCEEDINGS{ResNet101,
author = {K. He and X. Zhang and S. Ren and J. Sun},
booktitle = {Proceedings of the IEEE Conference on Computer Vision and Pattern Recognition},
title = {Deep Residual Learning for Image Recognition},
year = {2016},
volume = {},
  doi={10.1109/CVPR.2016.90},
pages = {770-778}}

@INPROCEEDINGS{MetaDet,
author = {Y. Wang and D. Ramanan and M. Hebert},
booktitle = {Proceedings of the IEEE International Conference on Computer Vision},
title = {Meta-Learning to Detect Rare Objects},
year = {2019},
volume = {},
doi={10.1109/ICCV.2019.01002},
pages = {9924-9933}}

@ARTICLE{Intelligence,
  author={J. Guo and W. Luo and  B. Song and F. Yu and X. Du},
  journal={IEEE Network}, 
  title={Intelligence-Sharing Vehicular Networks with Mobile Edge Computing and Spatiotemporal Knowledge Transfer}, 
  year={2020},
  volume={34},
  number={4},
  pages={256-262},
  doi={10.1109/MNET.001.1900512}}

@article{GUO202187,
title = {Trust-aware recommendation based on heterogeneous multi-relational graphs fusion},
journal = {Information Fusion},
volume = {74},
pages = {87-95},
year = {2021},
issn = {1566-2535},
doi = {https://doi.org/10.1016/j.inffus.2021.04.001},
url = {https://www.sciencedirect.com/science/article/pii/S1566253521000671},
author = {Jie Guo and Yan Zhou and Peng Zhang and Bin Song and Chen Chen}}

@INPROCEEDINGS{ONCE,
  author={Pérez-Rúa, Juan-Manuel and Zhu, Xiatian and Hospedales, Timothy M. and Xiang, Tao},
  booktitle={2020 IEEE/CVF Conference on Computer Vision and Pattern Recognition (CVPR)}, 
  title={Incremental Few-Shot Object Detection}, 
  year={2020},
  volume={},
  number={},
  pages={13843-13852},
  doi={10.1109/CVPR42600.2020.01386}}

@INPROCEEDINGS{FSOD,
  author={Fan, Qi and Zhuo, Wei and Tang, Chi-Keung and Tai, Yu-Wing},
  booktitle={2020 IEEE/CVF Conference on Computer Vision and Pattern Recognition (CVPR)}, 
  title={Few-Shot Object Detection With Attention-RPN and Multi-Relation Detector}, 
  year={2020},
  volume={},
  number={},
  pages={4012-4021},
  doi={10.1109/CVPR42600.2020.00407}}

@InProceedings{MPSR,
author="Wu, Jiaxi
and Liu, Songtao
and Huang, Di
and Wang, Yunhong",
title="Multi-scale Positive Sample Refinement for Few-Shot Object Detection",
booktitle="Computer Vision -- ECCV 2020",
year="2020",
publisher="Springer International Publishing",
pages="456--472"}
\end{document}